\documentclass{bmvc2k}
\usepackage{subfiles}
\usepackage[table]{xcolor}
\usepackage{tikz}
\usepackage{times}
\usepackage{epsfig}
\usepackage{graphicx}
\usepackage{amsmath}
\usepackage{amssymb}
\usepackage{adjustbox}
\usepackage{paralist}

\usepackage{pgfplots} 
\usepackage{pgfplotstable} 
\pgfplotsset{compat=newest} 
\usetikzlibrary{plotmarks} 
\usepackage[table]{xcolor}
\usepackage{booktabs} 

\title{Iterative Deep Learning for\newline Road Topology Extraction}

\addauthor{Carles Ventura}{cventuraroy@uoc.edu}{1}
\addauthor{Jordi Pont-Tuset}{jponttuset@vision.ee.ethz.ch}{2}
\addauthor{Sergi Caelles}{scaelles@vision.ee.ethz.ch}{2}
\addauthor{Kevis-Kokitsi Maninis}{kmaninis@vision.ee.ethz.ch}{2}
\addauthor{Luc Van Gool}{vangool@vision.ee.ethz.ch}{2}

\addinstitution{
 Scene Understanding and Artificial Intelligence Lab\\
 Universitat Oberta de Catalunya\\
 Barcelona, Spain
}
\addinstitution{
 Computer Vision Laboratory\\
 ETH Z\"{u}rich\\
 Z\"{u}rich, Switzerland
}

\runninghead{Ventura et al.}{Iterative Deep Learning for Road Topology Extraction}

\begin{document}

\maketitle

\begin{abstract}
This paper tackles the task of estimating the topology of road networks from aerial images.
Building on top of a global model that performs a dense semantical classification of the pixels of the image, we design a Convolutional Neural Network (CNN) that predicts the local connectivity among the central pixel of an input patch and its border points.
By iterating this local connectivity we sweep the whole image and infer the global topology of the road network, inspired by a human delineating a complex network with the tip of their finger. We perform an extensive and comprehensive qualitative and quantitative evaluation on the road network estimation task, and show that our method also generalizes well when moving to networks of retinal vessels.
\end{abstract}


\section{Introduction}
Deep learning has gone a long way since its jump to fame in the field of computer vision thanks to the outstanding results in the Imagenet~\cite{Rus+15} image classification competition back in 2012~\cite{Krizhevsky2012}.
We have witnessed the appearance of deeper~\cite{SiZi15} and deeper~\cite{He+16} architectures and the generalization to object detection with the well-known trilogy of R-CNNs~\cite{Gir+14, Gir15, Ren+15}.
Convolutional Neural Networks (CNNs) have played a central role in this development.

A significant step forward was done with the introduction of CNNs for dense prediction, in which the output of the system was not a classification of an image or bounding box into certain categories, but each pixel would receive an output decision.
The seminal fully convolutional network~\cite{LSD15} was able to perform per-pixel semantic segmentation thanks to an architecture without fully connected layers (i.e.\ fully convolutional).
Many tasks have been tackled from this perspective since then: semantic instance segmentation~\cite{Li+17, He+17}, edge detection~\cite{XiTu17}, medical image segmentation~\cite{maninis2016deep}, etc.

Other tasks, however, have a richer output structure beyond a per-pixel classification, and a higher abstraction of the result is expected.
Notable examples that have already been tackled by CNNs are the estimation of the human pose~\cite{NYD16}, or the room layout~\cite{Lee+17} from an image.
The common denominator of these tasks is that one expects an abstracted model of the result rather than a set of pixel classifications.

This work falls into this category by bringing the power of CNNs to the estimation of the \textbf{topology of filamentary networks} such as road networks from aerial images.
The structured output is of critical importance and priceless value in these applications: rather than knowing exactly which pixels in a aerial image are road or not, detecting whether two points are connected and how is arguably more informative.

\begin{figure}[t]
\centering
\includegraphics[width=0.96\linewidth]{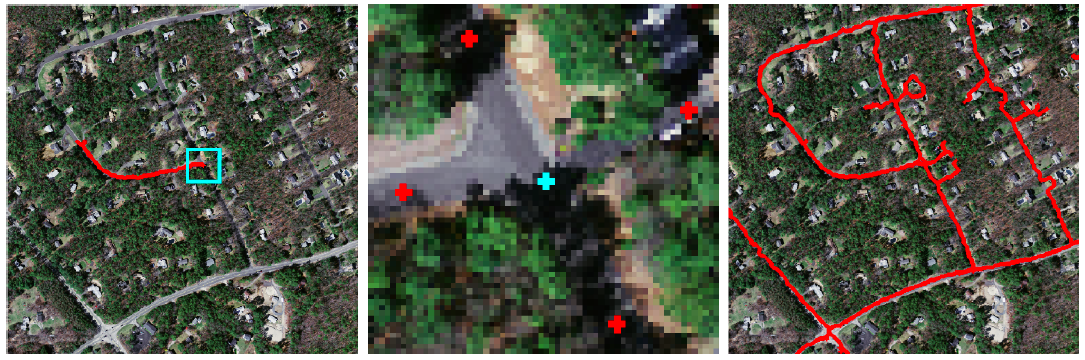}
\caption{Patch-based iterative approach for network topology extraction. \emph{Left}: Current state at some point of the iterative approach where the patch-level model for connectivity is applied over the
blue square. \emph{Center}: Detections at the local patch for the points at the border (in red) connected to the central point (in blue). \emph{Right}: Final result once the 
iterative approach ends.}
\label{fig:teaser-img}
\end{figure}

If one thinks how humans would extract the topology of an entangled graph network from an image, it might quickly come to mind the image of them tracing the filaments with the finger and \textit{sweeping} the connected paths continuously.
Inspired by this, we propose an iterative deep learning approach that sequentially connects dots within the filaments until it \textit{sweeps} all the visible network.
Our approach naturally allows incorporating human corrections: one can simply restart the tracing from the corrected point.

More specifically, we train a CNN on small patches that localizes input and exit points of the filaments within the patch (see image in the middle from Figure~\ref{fig:teaser-img}). 
By iteratively connecting these dots we obtain the global topology (graph) of the network (see right image from Figure~\ref{fig:teaser-img}).
We tackle the extraction of road networks from aerial photos.
We experiment on the publicly available Massachusetts Roads dataset to show that our algorithm improves over some strong baselines and provides accurate representations of the network topology. Code is available in \href{https://github.com/carlesventura/iterative-deep-learning}{https://github.com/carlesventura/iterative-deep-learning}.

\section{Related Work}
\paragraph{Curvilinear Structure Segmentation and Tracing:}
Tracing of curvilinear structures has been of broad interest in a range of applications, varying from blood vessel segmentation, roadmap segmentation, and reconstruction of human vasculature. Hessian-based methods rely on derivatives, to guide the development of a snake~\cite{WAC11}, or to detect vessel boundaries~\cite{Ban+12}. Model-based methods rely on strong assumptions about the geometric shapes of the filamentary structures~\cite{LaCh08, Soa+06} or are based on directional morphological operators~\cite{valero2010}. Learning-based methods emerged for the task, using support vector machines on line operators~\cite{RiPe07}, fully-connected CRFs~\cite{OrBl14}, gradient-boosting~\cite{Bec+13}, classification trees~\cite{GuCh15}, or nearest neighbours~\cite{SLF15}. Closer to our approach, the most recent methods rely on Fully Convolutional Neural Networks (FCNs), to segment retinal blood vessels~\cite{maninis2016deep, Fu+16} , or recover vascular boundaries~\cite{Mer+16}. Different than all the aforementioned method that result in binary structure maps, our method employs deep learning to trace the entire structure of the curvilinear structures, recovering their entire connectivity map. Also related to our method, the authors of~\cite{cheng2014tracing} trace blood vessels using directed graph theory. To the best of our knowledge, we are the first to apply deep learning for tracing curvilinear structures.

\paragraph{Road Centerline Detection:} 
Centerline detection has also followed the trend of curvilinear structure segmentation, with early attempts on gradient-based methods getting outperformed when stronger machine learning techniques emerged~\cite{WMS13, SLF14}. Sironi et al.~\cite{SLF15} model the relationships between neighbouring patches to reach the decision for the centerlines. Most recent works employ deep learning techniques, and include results on the Torontocity dataset~\cite{wang2017torontocity} (which has not been publicly released yet).~\cite{mattyus2017deep} is the most recent work on extracting the road topology from aerial images, and proposes a post processing 
algorithm that reasons about missing connections in the extracted road topology from an initial segmentation. In contrast, in our paper we propose an approach that learns the connectivity
of the roads at a local scale and is iteratively extended to the entire road network without relying on the results of an initial segmentation.

\section{Our Approach}

This section presents our approach, which combines a global scale for curvilinear structure segmentation and a local scale to estimate its connectivity. 
The current best approaches for curvilinear structure segmentation applies state-of-the-art deep learning techniques to obtain a segmentation map where each pixel is classified as 
belonging to the structure (foreground) or not (background).
Despite their good performance in segmentation evaluation measures,
one of the main drawbacks of these approaches is that they do not take any structure information into account.
In particular, these methods are blind to connectivity information among the points that lie in their predicted mask, since all points are assigned only a binary label.

\begin{figure}[t]
\centering
\includegraphics[width=0.80\linewidth]{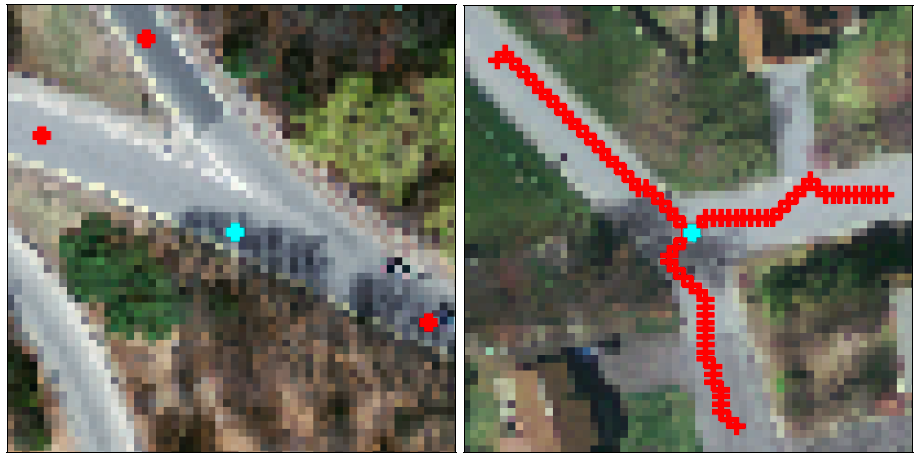}
\caption{\emph{Left}: Example of training patch for connectivity. The red points represent the locations from the patch border connected with the road indicated by the blue point in the center of the patch. \emph{Right}: Shortest path on semantic segmentation to connect the locations detected at the patch border with the patch center.}
\label{fig:training_patches}
\end{figure}

Section~\ref{sec:patch-level-connectivity} proposes a method that learns the connectivity of the elements at a local scale.
Given a patch of the image centered on a curvilinear structure, the model predicts the locations at the patch border connected with the centered structure.
Figure~\ref{fig:training_patches}-left shows some examples of how we formulate the local connectivity for aerial images: we learn to predict the points on the border of the patch (red dots) that are connected to the center pixel (blue dot). Once the local connectivity model is learned, it is iteratively applied to the image, connecting previous predictions with next ones, and gradually extracting the topology of the graph network, as explained in Section~\ref{sec:patch-level-iterative}. We present our evaluation metrics in Section~\ref{sec:evaluation}.


\subsection{Patch-level learning for connectivity}
\label{sec:patch-level-connectivity}
As introduced above, the goal is to train a model to estimate the local connectivity in patches. 
The concept of connectivity is not a property from single points but from pairs of pixels.
Current architectures, however, are designed to estimate per-pixel properties rather than pairwise information. 
To solve this issue, the local network is designed to estimate which points from the border in a patch are connected to a given input point.
Given a patch, therefore, we need to encode the position of the input and output points. 

In the context of human pose estimation~\cite{NYD16}, for instance, points have been encoded as heatmaps with Gaussians centered on them. We follow the same approach and thus the output of our model is a per-pixel probability of being a connected point.
Instead of encoding the input point by adding an extra input channel with a heatmap marking its position, we follow a simpler approach.
We always place the input point at the center of the patch, thus avoiding the extra input channel and further simplifying the model.
We see in the experiments that the model is indeed capable of learning that the central point is the input location we are interested in.

More precisely, we take the architecture of stacked hourglass networks~\cite{NYD16} (also used for human pose estimation) to learn the patch-based model for connectivity. This architecture is based on a repeated bottom-up, top-down processing used in conjunction with intermediate supervision. Each bottom-up, top-down processing block is referred to as an hourglass module, which is related to fully convolutional networks that process spatial information at multiple scales but with a more symmetric distribution.

The network is trained using a set of $k\!\times\!k$-pixel patches from the training set with the pixel at the center of the patch belonging to the foreground, e.g. a pixel annotated as road for road segmentation.
The output is a heatmap that predicts the probability of each location being connected to the central point of the patch.

We finally connect the border locations to the center locations by computing the shortest path through the semantic segmentation computed from the global model introduced before, as shown in Figure~\ref{fig:training_patches}-right. Note that the patch is local enough that a shortest path on the global model is reliable.


\subsection{Iterative delineation}
\label{sec:patch-level-iterative}
Once the patch-level model for connectivity has been learned, the model is applied iteratively through the image in order to extract the topology of the network, as a human delineating an image with its fingers not to lose the track.
We start from the point with highest foreground probability, given by the global model, as the starting point for the iterative sweeping approach.
We then center a patch on this point and find the set of locations at the border of the patch that are connected to the center, with their respective confidence values, using the local patch model.

We discard the locations with a confidence value below a certain threshold and add the remaining ones to a bag of points to be explored $B_E$.
For each predicted point, we store its location, its confidence value and its precedent predicted point (i.e. the point that was on the center of the patch when the point was predicted).
The predicted point $p$ from $B_E$ with the highest confidence value is removed from $B_E$ and inserted to a list of visited points $B_V$.

Then, $p$ is connected to its precedent predicted point using the Dijkstra~\cite{DIJKSTRA1959} algorithm over the segmentation probability map over the patch to find the minimum path between them.

We then iterate the process with a patch centered on $p_c$ and the new predicted points over the confidence threshold are appended to $B_E$ where they will \textit{compete} against the previous points in $B_E$ to be the next point to be explored.
This process is iteratively applied until $B_E$ is empty.
Note that the list of visited points $B_V$ is used to discard any point already explored and, therefore, to avoid revisiting the same points over and over again. In a patch centered on $p_c$, if a predicted point $p_p$ belongs to a local neighbourhood of a point $p_v \in B_V$ and $p_v$ is the precedent point of $p_c$, then the predicted point $p_p$ is discarded. Otherwise, if $p_v$ is not the precedent point of $p_c$ but $p_p$ belongs to a local neighbourhood of $p_v$, then the predicted point $p_p$ is considered to be connected with $p_c$, but $p_p$ will not be considered for expansion.

The algorithm has been generalized to tackle a problem with unconnected
areas, e.g. aerial road images may content roads that are not connected between them and, therefore, there could be roads not reachable from a single starting point. To prevent that some part of the network topology may have not been extracted, we select a new starting point for a new exploration once the previous $B_E$ is empty. We impose two constraints on the eligibility
for a new starting point: $(i)$ they have to be at a minimum distance of the areas already explored and $(ii)$ their confidence value on the segmentation probability map has to be over a 
minimum confidence threshold.
The iterative approach ends when there are no remaining points eligible for new starting points.

\subsection{Topology evaluation}
\label{sec:evaluation}
The output of our algorithm is a graph defining the topology of the input network, so we need metrics to evaluate their correctness.
We propose two different measures for this: a \textit{classical} precision-recall measure that evaluates which locations of the network are detected, and a metric to evaluate connectivity, by quantifying how many pairs of points are correctly or incorrectly connected.

To compute the classical precision-recall curve between two graphs, we build an image with a pixel-wide line sweeping all edges of the given graphs. We then apply the original precision-recall for boundaries \cite{martin2004learning} on these pair of images.
Precision $P$ refers to the ratio between the number of pixels correctly detected as boundary (true positives) and the number of pixels detected as boundary (true positives + false positives).
Recall $R$ refers to the ratio between the number of pixels correctly detected as boundary (true positives) and the number of pixels annotated as boundary in the ground truth (true positive + false negative).
We take the F measure between $P$ and $R$ as a trade-off metric.

The second measure is the connectivity $C$, inspired by the definition in~\cite{mattyus2017deep} as the ratio of segments which were estimated without discontinuities.
We define a segment in the graph as the curvilinear structure that connects two consecutive junctions in the ground-truth annotations, as well as connecting an endpoint 
and its closest connected junction. 
Two junctions are considered consecutive if there is no other junction within the line that connects them. Given the ground truth path between two consecutive junctions $p_{gt}$, the nearest point 
from the predicted network to each junction is retrieved. Then, the shortest path through the predicted network connecting the retrieved pair of points is computed, which is referred to as $p_{pred}$. The ratio between the length of $p_{gt}$ and the length of $p_{pred}$ is computed. If the ratio is greater than 0.8 we consider that the ground truth path $p_{gt}$ has been estimated without discontinuities.

We propose to also have an F measure that combines precision $P$ with connectivity $C$, the reason being that a high connectivity $C$ value alone does not imply the correctness of the vessels, as it could be a result where all pixels are connected to everything.
Adding a \textit{competing} precision measure prevents this from happening. 


For the rest of the paper, $F^R$ stands for the F measure computed with recall and precision for boundaries values, whereas
$F^C$ stands for the $F$ measure computed between connectivity and precision.

\section{Experiments}

The experiments for road topology extraction on aerial images have been carried out on the publicly available Massachusetts Roads Dataset~\cite{MnihThesis}, which includes 1108 images for training, 14
images for validation, and 49 images for testing. Each image is 1500$\times$1500 pixels in size, covering an area of 2.25 square kilometers.

\paragraph{Patch-level evaluation:}
To train the patch-level model for connectivity we randomly select 130 patches with size 64$\times$64 pixels from each image of the training set, all of them centered on any pixel annotated as road in the ground truth. The number of patches has been selected as a trade-off between having enough training data and diverse training data, i.e. avoiding very rendundant patches.
The ground-truth locations for the connectivity at the patch level are found by intersecting the skeletonized ground truth mask with a square of side $s$ pixels (slightly smaller than the patch size) centered on the patch.
The ground-truth output heatmap is then generated by adding some Gaussian peaks centered in the found locations. We obtain a precision value of 86.8\%, a recall value of 82.2\% and F=84.5\% in a set of 700 patches randomly cropped from the validation images.

\begin{figure}[t]
\centering
\includegraphics[width=\linewidth]{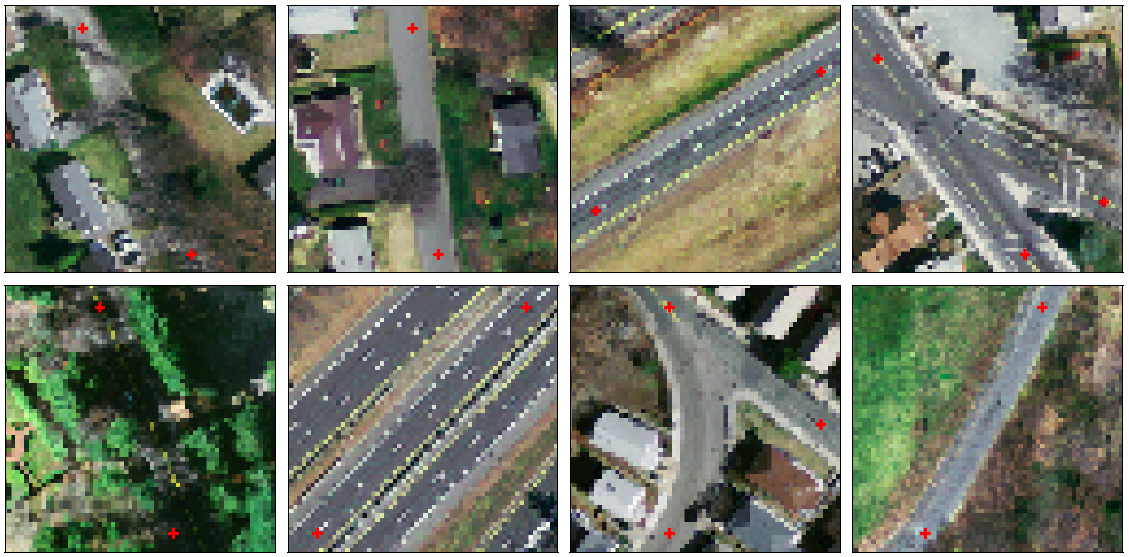} 
\caption{Visual results for the patch-level model for connectivity applied to patches from the Massachusetts Roads dataset. The red crosses represent the detections.}
\label{fig:patch-detector-roads}
\end{figure}

\begin{figure}[t]
\centering
\includegraphics[width=\linewidth]{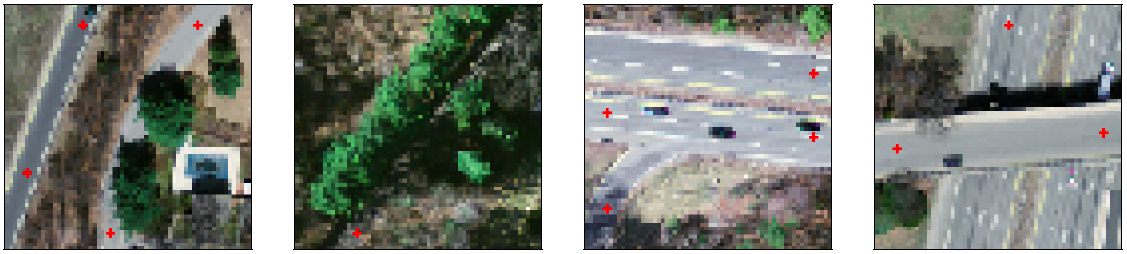}
\caption{Detection errors for the patch-level model for connectivity applied to patches from the Massachusetts Roads dataset. The red crosses represent the detections.}
\label{fig:patch-detector-roads-errors}
\end{figure}

Figure~\ref{fig:patch-detector-roads} shows some results for the patch-level model for connectivity applied to patches. We can see some examples where the road connections are found despite the shadows of the trees or the similarity of the background with the road. Furthermore, it also learns not to detect roads that are visible on the image but they are not connected to the central road.
Figure~\ref{fig:patch-detector-roads-errors} illustrates other examples where the model fails with false or missing detections, e.g. a visible road not connected with the central road has been wrongly detected.

\paragraph{Iterative delineation:}
Once the patch-level model for connectivity has been trained, it is iteratively applied to extract the topology of the roads from the aerial images.
As a baseline we compare to extracting the morphological skeleton of detections binarized at different thresholds from the architecture proposed in DRIU~\cite{maninis2016deep}, a VGG base network on which a set of specialized layers are trained to solve a retinal vessel segmentation task. This architecture has been analogously trained for the road segmentation task presented in this paper. Our proposed iterative approach uses this VGG-based architecture as the global model to select the starting point and to connect the points detected by the patch-level model with the central point of the patch (see Section~\ref{sec:patch-level-iterative}).

Table~\ref{tab:roads-results} shows the comparison between the global-based baseline for road segmentation based on a VGG architecture and our proposed iterative approach.
The connectivity of the skeleton resulting from the VGG-based road segmentation is very low, with a maximum precision-connectivity value $F^C = 49.3\%$ (a maximum precision-recall value $F^R=72.4\%$ is achieved with a different threshold). In contrast, our iterative delineation approach is able to reach a maximum precision-connectivity value $F^C = 74.4\%$, while also outperforming the classical precision-recall measure with $F^R = 81.6\%$.

\begin{table}
\centering
\begin{tabular}{ cccccc } 
 \toprule
 & P & R & C & $F^R$ & $F^C$\\ 
 \midrule
 VGG-150 & 49.2 & \bf{94.7} & 49.4 & 64.1 & 49.3\\
 VGG-175 & 61.5 & 88.6 & 30.7 & 72.0 & 41.0\\
 VGG-200 & 75.8 & 70.7 & 11.0 & 72.4 & 19.2\\
 Iterative (ours) & \bf{83.5} & 80.8 & \bf{67.1} & \bf{81.6} & \bf{74.4} \\
 \bottomrule
\end{tabular}
\vspace{1pt}
\caption{Boundary Precision-Recall and Connectivity evaluation in Massachusetts Roads dataset, where xxx in VGG-xxx refers to the threshold used on the output road segmentation from the VGG model before
extracting the skeleton.}
\label{tab:roads-results}
\end{table}

\begin{figure}
\centering
\includegraphics[width=\linewidth]{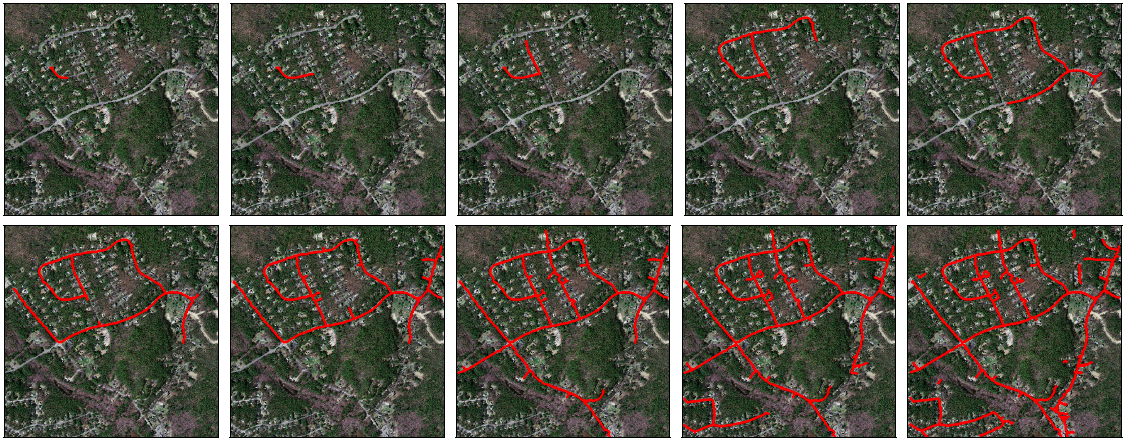}
\caption{Evolution of the road network in the iterative delineation. The progress is displayed from left to right and from top to bottom.}
\label{fig:progress-iterative-approach-roads}
\end{figure}

\begin{figure}[t]
\centering
\includegraphics[width=\linewidth]{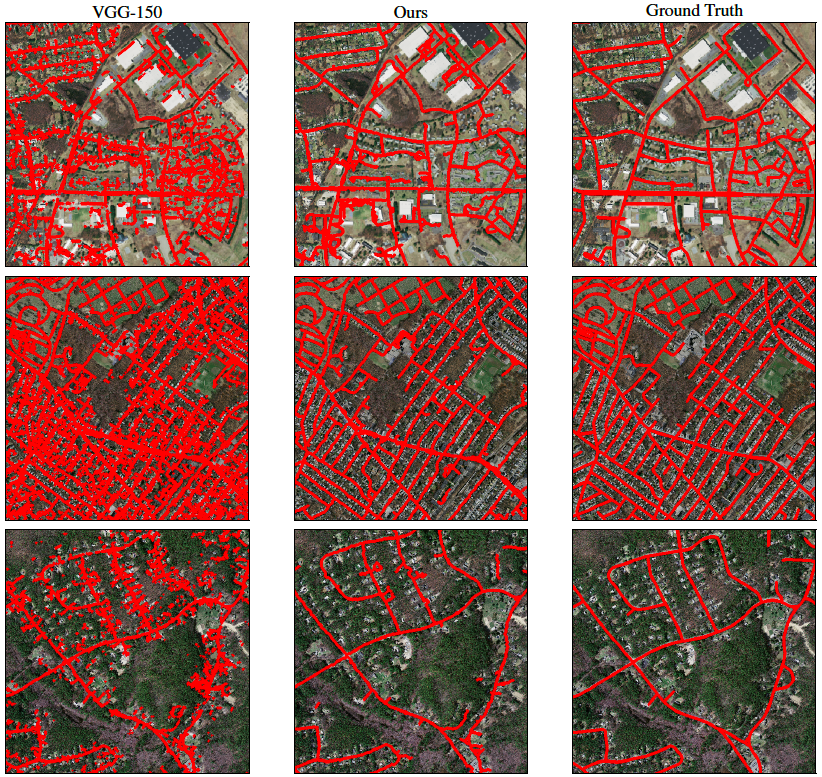} 
\caption{Results for road network topology extraction on the Massachusetts Roads dataset.
From left to right: VGG-150, our iterative delineation approach and ground truth.}
\label{fig:roads-results}
\end{figure}

\begin{figure}[t]
\centering
\includegraphics[width=\linewidth]{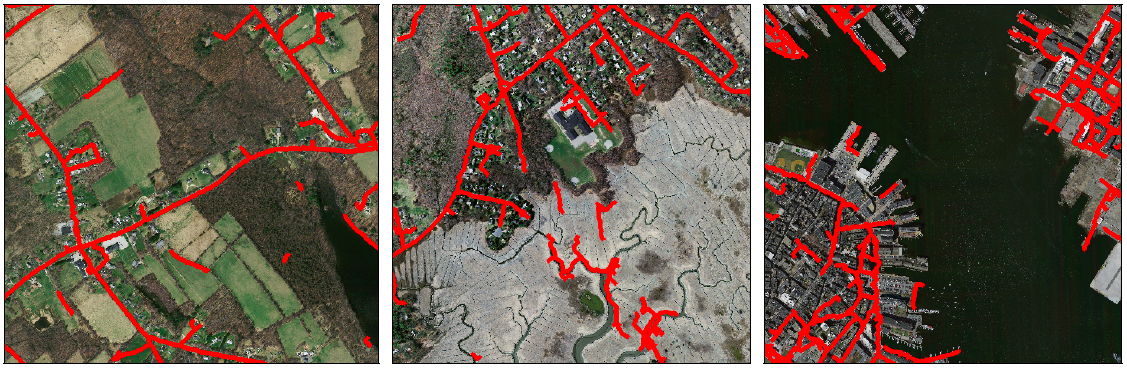} 
\caption{False and missing detections for road network topology extraction on the Massachusetts Roads dataset.}
\label{fig:roads-results-errors}
\end{figure}

Figure~\ref{fig:progress-iterative-approach-roads} illustrates how the road network topology delineation evolves along the iterations of our proposed approach for one of the test images.
Figure~\ref{fig:roads-results} shows some qualitative results in comparison with the VGG-based road segmentation baseline and the ground-truth annotations. Figure~\ref{fig:roads-results-errors}
shows some errors in the network topology delineation as some false detections on field (left image) and fluvial (central image) areas or missing detections on high density urban areas (right image).

\begin{figure}[t]
\centering
\includegraphics[width=0.96\linewidth]{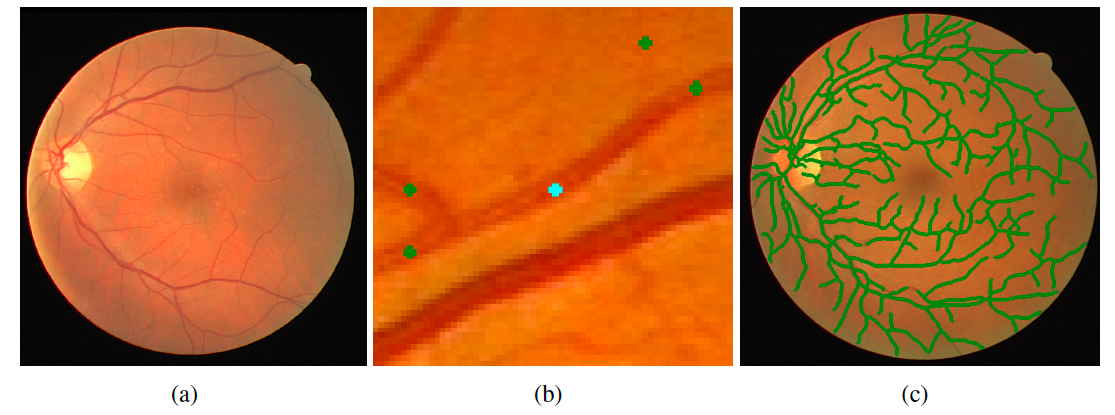}
\caption{Patch-based iterative approach for vessel network topology extraction. (a): Input retinal image. (b): Detections at the local patch for the points at the border (in green) connected to the central point (in blue). (c): Final result once the iterative approach ends.}
\label{fig:vessels-img}
\end{figure}

\begin{figure}[t]
\centering
\includegraphics[width=\linewidth]{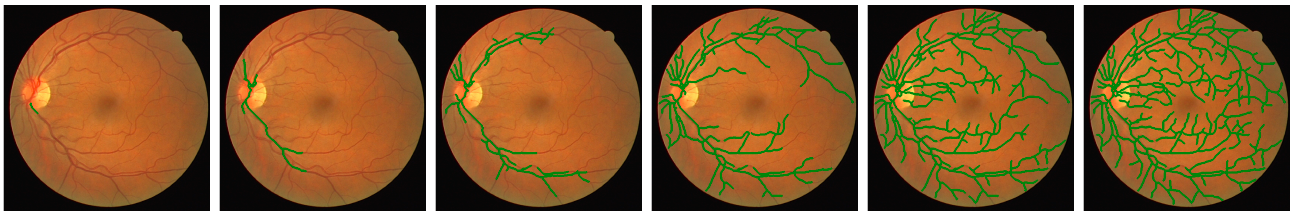}
\caption{Evolution of the vessel network in the iterative delineation.}
\label{fig:vessels-iterative}
\end{figure}

Finally, some experiments have also been performed on vessel segmentation from retinal images to show that the proposed approach is also valid in other fields. Experiments performed on the DRIVE~\cite{staal04ridge} dataset obtained similar precision-recall values in comparison with DRIU~\cite{maninis2016deep}, which is the state-of-the-art, but significatively higher connectivity values (see Table~\ref{tab:vessels-results}). Figure~\ref{fig:vessels-img} shows how the model has been trained and Figure~\ref{fig:vessels-iterative} shows the evolution of the iterative approach on a retinal image.

\begin{table}[!h]
\centering
\begin{tabular}{cccccc} 
 \toprule
 & P & R & C & $F^R$ & $F^C$\\ 
 \midrule
 DRIU \cite{maninis2016deep} & \bf{97.3} & 84.7 & 67.7 & \bf{90.4} & 79.8\\
 Iterative (ours) & 86.1 & \bf{94.1} & \bf{84.9} & 89.8 & \bf{85.5}\\
 \bottomrule
\end{tabular}
\vspace{1mm}
\caption{Quantitative results on DRIVE dataset for vessel segmentation.}
\label{tab:vessels-results}
\end{table}

\section{Conclusions}

In this paper we have presented an approach that iteratively applies a patch-based CNN model for connectivity to extract the topology of filamentary networks.
We have demonstrated the effectiveness of our technique on road networks from aerial photos.
The patch-based model is capable of learning that the central point is
the input location and of finding the locations at the patch border connected to the center.

A new  F measure ($F^C$) that combines precision and connectivity has been proposed to evaluate the topology results.  The experiments carried out on aerial images have obtained the best performance on $F^C$ compared to strong baselines.

Finally, the proposed iterative patch-based model has also been validated on vessel network extraction from retinal images to show the robustness of the approach in another field.

\section*{Acknowledgements}

This research was supported by the Spanish Ministry of Economy and Competitiveness (TIN2015-66951-C2-2-R grant), by Swiss Commission for Technology and Innovation (CTI, Grant No. 19015.1 PFES-ES, NeGeVA) and by the Universitat Oberta de Catalunya.

\clearpage

\bibliography{egbib}
\end{document}